\documentclass[runningheads]{llncs}

\usepackage{eccv}

\usepackage{eccvabbrv}

\usepackage{graphicx}
\usepackage{booktabs}

\usepackage[accsupp]{axessibility}  

\usepackage{enumitem}
\usepackage{graphicx}
\usepackage{wrapfig}
\usepackage{multirow}
\usepackage{makecell}
\usepackage{caption}
\usepackage{algorithm}
\usepackage{bm}
\usepackage{arydshln}
\usepackage{float}
\usepackage{subcaption}
\usepackage{amsmath}
\usepackage{booktabs}
\usepackage{arydshln}
\usepackage{amsmath}
\definecolor{mydarkgreen}{RGB}{0,115,0}

\usepackage{graphicx}
\usepackage{xcolor}
\usepackage{amsmath,amssymb}
\usepackage{algpseudocodex}

\definecolor{darkgreen}{rgb}{0.0,0.45,0.0}

\usepackage{hyperref}

\usepackage{orcidlink}

\begin{document}

\title{PointLAM: Local Attentive Mamba for Efficient Point-based 3D Object Detection} 

\titlerunning{PointLAM}

\author{Xuanming Shang \and
Weijia Zhang \and
Chao Ma\thanks{Corresponding author.}}

\authorrunning{X. Shang et al.}

\institute{
MoE Key Lab of Artificial Intelligence, Institute of AI, Shanghai Jiao Tong University, Shanghai, China\\
\email{\{sxm2021,weijia.zhang,chaoma\}@sjtu.edu.cn}
}
\maketitle

\begin{abstract}
3D object detection from LiDAR point clouds faces a fundamental dilemma: voxel-based methods achieve efficiency at the cost of geometric quantization, while point-based methods preserve fidelity but suffer from prohibitive computational bottlenecks. Specifically, point-based architectures are crippled by slow downsampling strategies (\eg, FPS) and expensive dynamic neighbor queries (\eg, $k$-NN) coupled with costly continuous interactions.
To tackle these systemic inefficiencies, we propose PointLAM, a highly efficient and powerful point-based architecture driven by two synergistic innovations. First, to resolve the downsampling bottleneck, we develop the Laplacian Point Sampler (LPS). LPS employs an implicit discrete Laplacian high-pass filter and Doubly Sorted Sampling to achieve fast, structure-aware foreground preservation. Second, to overcome local modeling latency, we design the Local Hadamard Aggregator (LHA). LHA decouples spatial indexing from feature representation using transient grids, and replaces complex continuous interactions with a Hadamard Gating mechanism for topology-aware, attentive modulation. By coupling this local gating with Bi-Directional Mamba (BDM) layers for global sequence modeling, we formulate the Local Attentive Mamba (LAM) block. Powered by this architecture, PointLAM achieves competitive performance on nuScenes and Waymo for point-based detectors. It rivals highly optimized voxel competitors while requiring a fraction of the computational footprint, demonstrating marked superiority in detecting small instances and handling extreme sparsity. Project page: https://pointlam.github.io/.

\keywords{3D Object Detection \and Point Cloud \and Mamba \and Autonomous Driving}

\end{abstract}

\section{Introduction}

\begin{figure}[t]  
    \centering  
    \includegraphics[width=\linewidth]{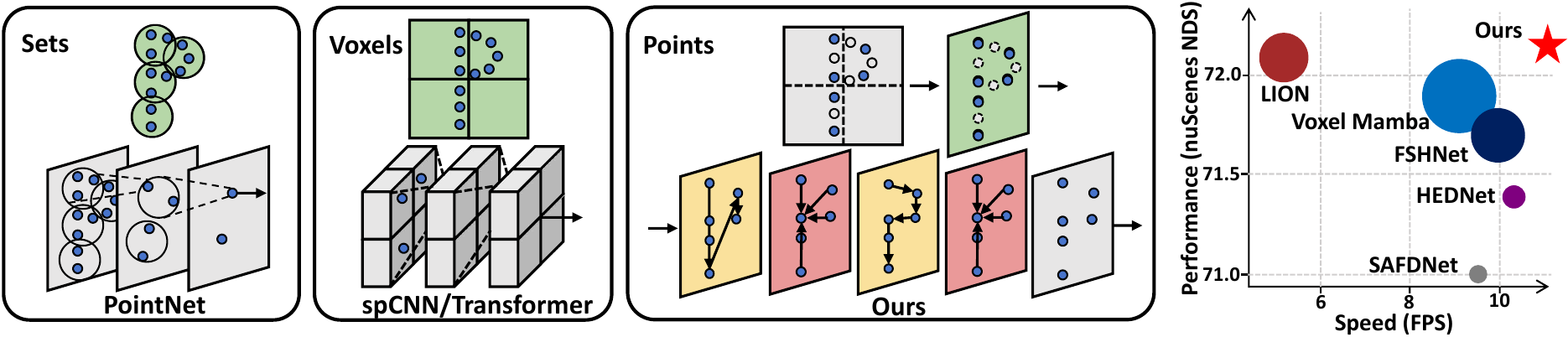} 
    \caption{Comparison of 3D point cloud processing paradigms. \textbf{Left:} Point-based methods~\cite{qi2017pointnet, qi2017pointnetpp, shi2019pointrcnn, yang20203dssd} rely on expensive neighborhood queries, while voxel-based methods~\cite{liu2024lion, zhang2023hednet, zhang2024safdnet, wang2023dsvt} sacrifice fine geometric details for efficiency. \textbf{Right:} PointLAM offers a favorable performance-efficiency trade-off on nuScenes. It is faster and lighter (marker size indicates parameter count) than several strong voxel-based baselines (circles), positioning it as a competitive point-based model (red star).}  
    \label{fig:point-based}  
\end{figure} 

Accurate 3D object detection from LiDAR point clouds, a cornerstone of modern autonomous systems, hinges on effectively processing vast and unstructured raw data. Historically, the field has confronted a fundamental dilemma between geometric fidelity and computational efficiency, diverging into voxel-based and point-based paradigms.
While voxel-based approaches efficiently process structured volumetric grids via 3D Sparse CNNs~\cite{fan2024fsd, zhang2024safdnet, zhang2023hednet, chen2023voxelnext, chen2023largekernel3d, shi2023pv, fan2022fully, yin2021center, shi2020points, liu2025fshnet} or Transformers~\cite{wu2024pointv3, liu2024seed, wang2023dsvt, liu2023flatformer, bai2022transfusion, zhou2022centerformer, mao2021voxel}, they inherently sacrifice geometric fidelity. In contrast, point-based methods~\cite{shi2019pointrcnn, yang20203dssd, zhang2022not, shi2020point} operate directly on raw point sets, preserving precise, fine-grained geometric details and avoiding quantization errors and information loss inherent to voxelization. As summarized in Fig.~\ref{fig:point-based}, these paradigms occupy different performance-efficiency regimes, motivating PointLAM as a point-level design for a better accuracy-efficiency balance.

However, systemic inefficiency hinders the widespread adoption of point-based detectors, critically manifesting in two fundamental bottlenecks.
First, the downsampling bottleneck restricts foreground preservation. The dominant Farthest Point Sampling (FPS)~\cite{yang20203dssd, liang2024pointmamba, han2024mamba3d, shi2020pv, shi2023pv} is computationally prohibitive and structure-agnostic, whereas semantic-dependent alternatives~\cite{yang20203dssd, zhang2022not} incur heavy overhead and lack robust geometric priors.
Second, capturing intricate local geometries mandates expensive spatial queries (\eg, $k$-NN) and heavy continuous interactions~\cite{qi2017pointnet, qi2017pointnetpp, shi2019pointrcnn, yang20203dssd}, severely degrading latency.

To tackle these issues, we introduce PointLAM, an efficient point-based architecture driven by two synergistic innovations. For the downsampling bottleneck, we propose the Laplacian Point Sampler (LPS). Instead of relying on structure-blurring local pooling~\cite{lang2019pointpillars, zhou2018voxelnet} or computation-heavy semantic samplers~\cite{zhang2022not}, LPS exploits a fundamental high-frequency geometric prior. Specifically, by calculating the feature deviation from the local neighborhood mean, we construct an implicit discrete Laplacian high-pass filter~\cite{wardetzky2007discrete} that inherently highlights distinct structural boundaries and corners. Guided by these Laplacian saliency scores, our Doubly Sorted Sampling (DSS) module ranks and selects points globally and regionally, ensuring uniform spatial coverage while strictly preserving informative geometric skeletons without the latency of FPS.

For the local modeling bottleneck, we propose the Local Hadamard Aggregator (LHA). Traditional queries like $k$-NN are computationally prohibitive and density-sensitive, often introducing severe spatial noise by retrieving distant irrelevant points in sparse regions~\cite{wang2024pointattn}. LHA resolves this by utilizing a transient grid strictly as a deterministic spatial router. This imposes a bounded receptive field to isolate distant noise, allowing sparse convolutions to aggregate context directly over continuous points. Furthermore, LHA replaces computation-heavy continuous relative positional embeddings~\cite{wu2022point} with a Hadamard Gating mechanism, adaptively modulating channel-wise amplitudes via the Hadamard product between a point's intrinsic feature and its aggregated context.

To structurally unify local and global modeling, we formulate the Local Attentive Mamba (LAM) block. While standard Mamba models~\cite{gu2023mamba, zhu2024vision} offer linear-time sequence modeling for long sequences, their 1D nature disrupts spatial structures when directly applied to 3D point clouds, degrading fine-grained local geometries. The LAM block resolves this limitation by coupling LHA with Bi-Directional Mamba (BDM) layers. Within this block, LHA functions as a topology-aware attentive gating mechanism that anchors local geometric priors before serialization, while BDM processes the serialized streams to establish global dependencies. This synergy allows PointLAM to deliver linear-complexity global perception while strictly preserving local geometric fidelity.

Extensive empirical evaluations systematically corroborate our theoretical formulations. Evaluated on the nuScenes~\cite{caesar2020nuscenes} and Waymo Open Datasets~\cite{sun2020scalability}, PointLAM achieves competitive accuracy on both benchmarks while maintaining a point-based architecture, approaching or matching several strong voxel-based baselines. Crucially, these competitive accuracies are achieved while requiring a fraction of the computational footprint—drastically reducing parameters, FLOPs, and inference latency. Furthermore, PointLAM demonstrates strong performance in detecting small instances and handling extreme sparsity, directly validating the geometric fidelity preserved by our point-based architecture.

\section{Related Work}
\label{sec:related_work}

\subsection{LiDAR-based 3D Object Detection}

LiDAR-based 3D object detectors can be broadly categorized into two types: point-based and voxel-based methods. Point-based methods~\cite{qi2017pointnet, qi2017pointnetpp, shi2019pointrcnn, yang20203dssd, zhang2022not, shi2020point} directly process raw, irregular point sets using techniques such as the PointNet series~\cite{qi2017pointnet, qi2017pointnetpp} to extract geometric features from local point neighborhoods. While preserving precise location information, these methods often contend with challenges such as high computational costs for sampling and grouping, lower inference efficiency, and difficulties in capturing expansive contextual features due to their localized processing. 
In contrast, voxel-based approaches~\cite{wang2023dsvt, fan2022embracing, bai2022transfusion, chen2023voxelnext, zhang2023hednet, jin2025unimamba, liu2025fshnet} convert unstructured point clouds into regular 3D voxel grids. This regularization allows the application of more conventional network architectures and has led to voxel-based methods becoming a mainstream approach in 3D object detection.

Voxel-based methods can be broadly categorized into Sparse Convolutional Neural Network (SpCNN)-based and Transformer-based. SpCNN-based methods~\cite{yin2021center, chen2023voxelnext, li2023pillarnext, fan2024fsd, zhang2024safdnet, zhang2023hednet, chen2023largekernel3d, shi2020points, yan2018second, deng2021voxel, fan2022fully} utilize 3D sparse convolutions, which efficiently process only non-empty voxels. While efficient for sparse data, the reliance on small kernels can restrict the effective receptive field, limiting the capture of long-range dependencies. Transformer-based methods~\cite{bai2022transfusion, fan2022embracing, wang2023dsvt, liu2024seed, dong2022mssvt, li2023mssvt, sun2022swformer, zhou2022centerformer, liu2023flatformer, wu2024pointv3, zhu2023conquer, mao2021voxel, he2022voxel, liu2025fshnet} have been introduced to the voxel domain, grouping voxels and applying self-attention to model global relationships. However, the quadratic complexity of Transformers often necessitates processing a limited number of voxels or voxel groups to remain computationally feasible. 
The constraints of restricted receptive fields in SpCNNs and the computational demands or limited scope of voxel-based Transformers motivate the exploration of alternative architectures like Mamba, which PointLAM leverages within its point-based framework to capture global context efficiently.

\subsection{Mamba for 3D Point Cloud Processing}

Mamba~\cite{gu2023mamba} has recently been introduced into deep neural networks as a compelling alternative to Transformers. Mamba incorporates input-dependent parameters and a selection mechanism to achieve linear-time sequence modeling and strong performance. Its success has spurred adaptations to general vision tasks such as image classification, semantic segmentation, and 2D object detection, with models such as Vision Mamba~\cite{zhu2024vision} and Vmamba~\cite{liu2024vmamba} employing different bidirectional SSMs or 2D-selective scanning techniques to process image data and learn global visual clues effectively.

Building on these advancements, leveraging Mamba for 3D point cloud analysis is an emerging research frontier, initiated by pioneering efforts such as PointMamba~\cite{liang2024pointmamba}. Applying these inherently sequential models to unordered and unstructured 3D point clouds presents distinct challenges: devising effective point serialization strategies that preserve spatial locality, adequately capturing fine-grained local geometric details (which is not the primary strength of global-centric Mamba), and addressing the non-causal nature of 3D spatial relationships. Subsequent works like Mamba3D~\cite{han2024mamba3d}, Point cloud mamba~\cite{zhang2025point}, and others (\textit{e.g.}, \cite{liu2024lion, zhang2024voxel, jin2025unimamba}) have explored various solutions, often involving specific serialization protocols, such as coordinate sorting \cite{liu2024vmamba} and space-filling curves \cite{sagan1993three, orenstein1986spatial}, specialized scanning methods, or the integration of components such as convolutions \cite{liu2024lion, zhang2024voxel, jin2025unimamba} to boost local feature extraction.

\section{Method}
\label{sec:method}

This section details the architecture of PointLAM, our proposed efficient point-based 3D detector. We begin with an overview of the overall framework (Sec. \ref{sec:overall}), followed by in-depth descriptions of its two core innovations: the Laplacian Point Sampler (Sec. \ref{sec:sample}) and the PointLAM backbone block (Sec. \ref{sec:block}).

\subsection{Overall Architecture}
\label{sec:overall}

\begin{figure}[t]
    \centering
    \includegraphics[width=0.90\linewidth]{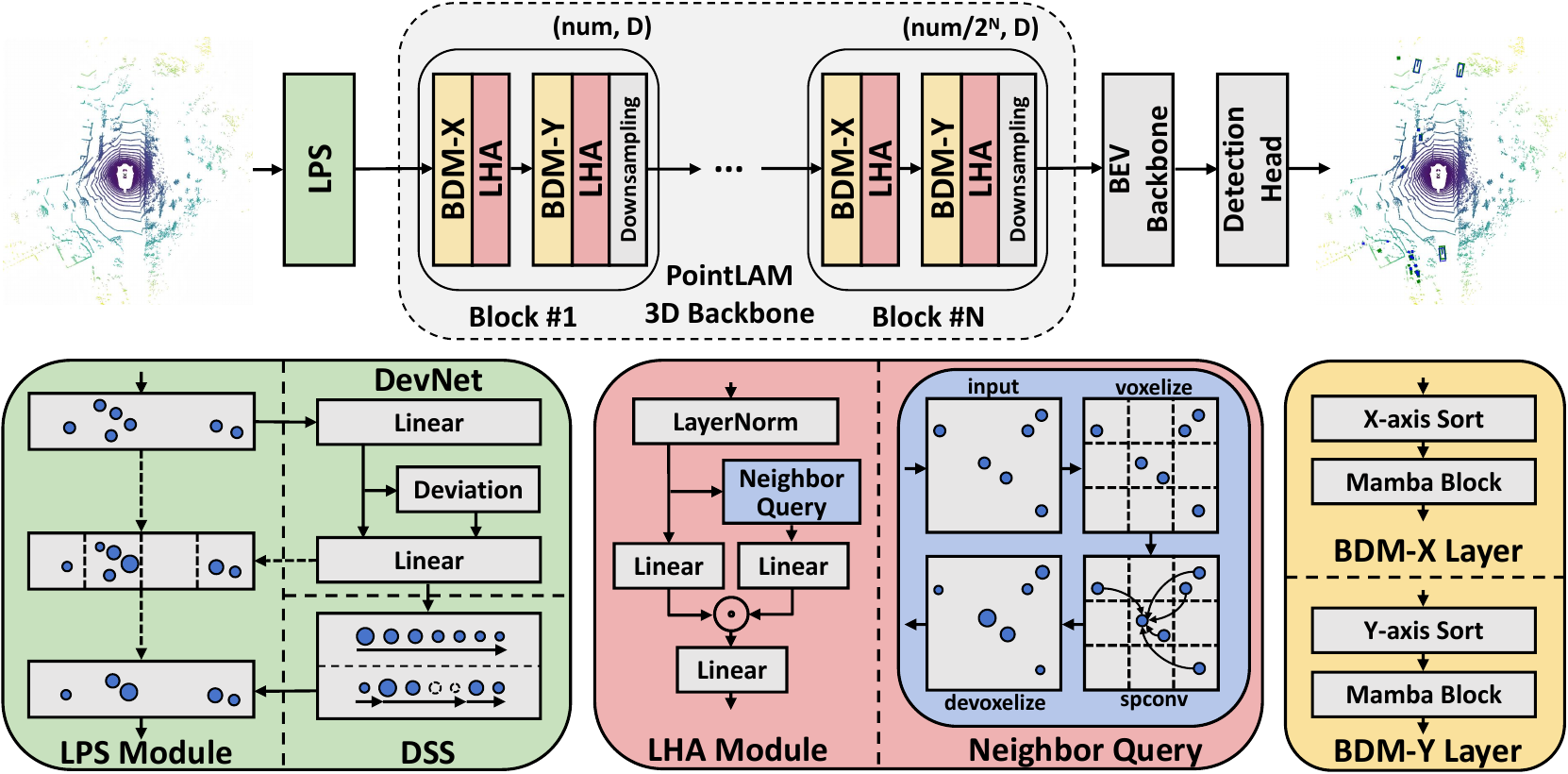} 
\caption{The overall architecture of PointLAM. \textbf{Top:} The pipeline starts with the Laplacian Point Sampler (LPS) for fast, structure-aware downsampling, followed by a 3D backbone of $N$ blocks. Each block cascades X-axis or Y-axis Bi-Directional Mamba (BDM) layers for global context, paired with a Local Hadamard Aggregator (LHA) module for efficient local topology modeling, followed by BEV projection and detection heads.
\textbf{Bottom-Left:} LPS utilizes a Deviation Network (DevNet) to effectively implement an implicit discrete Laplacian high-pass filter, combined with Doubly Sorted Sampling (DSS) to strictly preserve informative foreground structures.
\textbf{Bottom-Center:} LHA captures local geometry by spatially indexing via a transient voxel grid, and employs a Hadamard Gating mechanism for topology-aware feature modulation.
\textbf{Bottom-Right:} BDM establishes global context by sequentially processing points sorted along the X and Y axes via Mamba blocks.}
    \label{fig:Backbone}
\end{figure}

As illustrated in Fig.~\ref{fig:Backbone}, PointLAM processes raw point clouds through a streamlined pipeline comprising the Laplacian Point Sampler (LPS) and a hierarchical 3D backbone. 
First, the proposed LPS (Sec.~\ref{sec:sample}) efficiently downsamples the massive input. It utilizes a Deviation Network (DevNet) to explicitly encode the geometric distinctiveness of each point. This is followed by a Doubly Sorted Sampling (DSS) strategy to rapidly and strictly retain the most informative foreground structures.
The sampled points are then processed by the 3D backbone, which consists of a stack of $N$ Local Attentive Mamba (LAM) blocks. To synergize global and local modeling, each block sequentially integrates Bi-Directional Mamba (BDM) layers (Sec.~\ref{sec:bdm}) for long-range dependency modeling, and Local Hadamard Aggregator (LHA) modules (Sec.~\ref{sec:lma}) for fine-grained feature extraction.
Notably, LHA circumvents the severe computational overhead of dynamic neighbor queries (\eg, k-NN) by utilizing a transient grid to decouple spatial indexing and isolate distant noise. This is followed by a Hadamard Gating mechanism to adaptively modulate features based on local topology.
Finally, after stage-wise downsampling, the point features are projected into a Bird's-Eye-View (BEV) representation for the detection head.

\subsection{Laplacian Point Sampler}
\label{sec:sample}

Point-based 3D detectors critically depend on downsampling to reduce massive input clouds to manageable scales. However, the dominant Farthest Point Sampling (FPS) \cite{yang20203dssd, liang2024pointmamba, han2024mamba3d, shi2020pv, shi2023pv} imposes a prohibitive computational bottleneck due to its iterative nature and remains inherently agnostic to foreground structures. Conversely, recent semantic-aware samplers incur heavy computational overhead for feature extraction and lack robust geometric priors~\cite{yang20203dssd, zhang2022not}. To resolve this fundamental inefficiency, we introduce the Laplacian Point Sampler (LPS), a highly efficient two-stage pipeline driven by a high-frequency geometric prior. The LPS operates via a Deviation Network (DevNet) to explicitly encode geometric distinctiveness, followed by a Doubly Sorted Sampling (DSS) algorithm for fast, structure-aware point preservation.

\noindent\textbf{Deviation Network.}
Given a partitioned point cloud, we project each point $p_i$ and its local offsets into an intermediate feature $h_i \in \mathbb{R}^D$. Standard Point Feature Networks (PFNs) typically aggregate these using max-pooling. However, max-pooling acts as a spatial low-pass filter with a sparse, ``Winner-Take-All'' gradient flow ($\frac{\partial \mathcal{L}}{\partial h_{\text{non-max}}} = 0$). Consequently, structurally critical but numerically weaker foreground signals are severely suppressed by dominant background features.

To resolve this, DevNet replaces max-pooling with a Deviation operation, computing the difference between each point's feature $h_i$ and its local spatial mean $\bar{h}$:
\begin{equation}
    \delta_i = h_i - \bar{h} = \frac{1}{N} \sum_{j=1}^N (h_i - h_j)
\end{equation}
Mathematically, this difference exactly formulates the discrete Laplace operator. As a spatial derivative measuring local signal roughness, the Laplacian yields near-zero responses on smooth, low-frequency surfaces ($h_i \approx \bar{h}$) and high-magnitude responses at high-frequency geometric singularities (e.g., corners, boundaries). DevNet thus functions as an explicit high-pass filter, robustly preserving geometrically distinctive points.

Crucially, this Laplacian formulation guarantees a dense gradient flow. During backpropagation, the gradient for any point $p_i$ is derived as:
\begin{equation}
    \frac{\partial \mathcal{L}}{\partial h_i} = \frac{\partial \mathcal{L}}{\partial \delta_i} \left( 1 - \frac{1}{N} \right) - \frac{1}{N} \sum_{j \ne i} \frac{\partial \mathcal{L}}{\partial \delta_j}
\end{equation}
Unlike the zero-sum nature of max-pooling, this continuous formulation drives the network to learn cooperative contrast. Every point receives active gradient updates, optimizing the network to maximize the statistical distinctiveness of salient points ($\delta_i$) relative to their neighborhood ($\delta_j$). The final augmented feature is then constructed by concatenating the intrinsic feature with its Laplacian response: $F'_{p_i} = \text{Concat}(h_i, \delta_i)$.

\noindent\textbf{Doubly Sorted Sampling.}
Guided by the Laplacian responses encoded in $F'_{p_i}$, the DSS algorithm performs fast, structure-aware point selection. First, an importance score $S(p_i) = ||F'_{p_i}||_2$ is derived for each point $p_i$, directly reflecting its distinctiveness. 

To entirely circumvent the prohibitive latency of iterative Farthest Point Sampling (FPS), DSS replaces sequential distance queries with an efficient two-step sorting mechanism. Points are first globally sorted by $S(p_i)$ in descending order, and then stably sorted by their region indices. This ensures the most salient geometric points lead each regional subset. Let $p'_j$ denote the $j$-th point in this doubly sorted list, with region index $r'_j$. The regional top-$k$ selection rule is strictly formulated as:
\begin{equation}
\text{Select } p'_j \iff \begin{cases} \text{True} & \text{if } j < k \\ r'_j \neq r'_{j-k} & \text{if } j \ge k \end{cases}
\end{equation}
Essentially, this deterministic rule selects the first $k$ points globally ($j<k$) and subsequently retains a point only if its region index differs from the $k$-th preceding point ($r'_j \neq r'_{j-k}$). Consequently, the DSS strategy guarantees uniform spatial coverage while strictly preserving the informative geometric skeletons of foreground objects. By reducing the downsampling bottleneck to a simple sorting operation, it preserves essential local structures for the subsequent PointLAM blocks at a fraction of the computational cost.

\subsection{PointLAM Block}
\label{sec:block}

The Local Attentive Mamba (LAM) block constitutes our PointLAM feature extraction pipeline. It synergizes fine-grained local attentive modulation with linear-complexity global sequence modeling to efficiently process unstructured point clouds. This section details its two structurally unified sub-modules: the Local Hadamard Aggregator (LHA) for topology-aware local gating, and the Bi-Directional Mamba (BDM) for global context perception.

\subsubsection{Local Hadamard Aggregator (LHA)}
\label{sec:lma}

Capturing fine-grained local geometry is vital for 3D detection. However, traditional point-based methods \cite{qi2017pointnet, qi2017pointnetpp, shi2019pointrcnn, yang20203dssd} relying on explicit continuous neighborhood queries (e.g., k-NN) are computationally prohibitive and density-sensitive, often introducing severe spatial noise by retrieving distant irrelevant points in sparse regions. To resolve this local modeling bottleneck, we introduce the Local Hadamard Aggregator (LHA).

\noindent\textbf{Neighbor Query.}
To circumvent the latency of dynamic queries, LHA decouples spatial indexing from feature representation by utilizing a transient grid. For each point $p_i$ with coordinates $\mathbf{p}_i$ and features $\mathbf{f}_i \in \mathbb{R}^D$, we associate it with a discrete integer grid index $\mathbf{v}_i = \mathcal{V}(\mathbf{p}_i)$. Acting purely as a deterministic router rather than a quantization step, this grid imposes a bounded receptive field to strictly isolate distant noise. The neighborhood $\mathcal{N}(i)$ is defined as the set of points whose grid indices fall within a fixed kernel window (e.g., $3\times3\times3$) centered on $\mathbf{v}_i$. Local aggregation is then performed via a learnable sparse convolutional kernel $W$:
\begin{equation}
\mathbf{f}^{\text{local}}_i = \sum_{j \in \mathcal{N}(i)} W(\mathcal{V}(\mathbf{p}_j) - \mathcal{V}(\mathbf{p}_i)) \cdot \mathbf{f}_j
\end{equation}\label{eq:lma_conv}
Crucially, point features are never merged or averaged into a voxel representation. The grid serves purely as a structural scaffold, allowing sparse convolutions to aggregate context directly over individual, continuous points. The resulting context-rich features $\{\mathbf{f}^{\text{local}}_i\}$ are then passed through a linear layer to produce a local context representation $\mathbf{c}_i \in \mathbb{R}^D$.

\noindent\textbf{Topology-Aware Hadamard Gating.}
To adaptively refine the aggregated local features, LHA abandons computation-heavy neighbor interactions—such as continuous relative positional embeddings—in favor of a Hadamard Gating mechanism. The original input feature $\mathbf{f}_i$ is passed through a parallel linear layer to produce an intrinsic gating vector $\mathbf{g}_i \in \mathbb{R}^D$. 

By simply computing the Hadamard product ($\odot$) between this gating vector and the aggregated local context $\mathbf{c}_i$, LHA adaptively modulates channel-wise amplitudes based on local topology. Stabilized by a residual connection, the complete LHA operation is formulated as:
\begin{equation}
    \mathbf{f}^{\text{out}}_i = (\mathbf{g}_i \odot \mathbf{c}_i) + \mathbf{f}_i
\end{equation}
This bipartite design is simple and highly efficient. The transient grid secures rich local geometry without dynamic spatial searches, while the Hadamard gating provides a powerful, topology-aware feature refinement with negligible computational overhead.

\subsubsection{Bi-Directional Mamba and LAM Synergy}
\label{sec:bdm}

While the Local Hadamard Aggregator (LHA) establishes robust local geometric priors, perceiving 3D structures necessitates global context. Standard Mamba models excel at linear-time sequence modeling, effectively overcoming the quadratic complexity of self-attention for the ultra-long point sequences typical of outdoor LiDAR scenes. However, as inherently 1D sequence models, vanilla Mamba architectures inevitably disrupt the spatial locality of unstructured 3D point clouds during serialization. 

The Local Attentive Mamba (LAM) block resolves this fundamental limitation by structurally coupling LHA and Bi-Directional Mamba (BDM) in a synergistic cascade. Rather than enforcing a strict precedence, LAM interleaves global scanning with local attentive modulation. As BDM processes the serialized streams to establish long-distance dependencies, the accompanying LHA modules act as topological anchors. They continually restore and reinforce fine-grained geometric fidelity, effectively preventing the spatial disruption inherent to 1D serialization from compounding across deep layers.

To capture this global context, BDM serializes the unstructured point sets. As illustrated in Fig.~\ref{fig:Backbone}, we deliberately adopt a simple axis scan, merging all samples within a batch into a single continuous sequence ordered along the X-axis or Y-axis. The X-sorted points are processed through a Mamba block to capture X-direction dependencies. The intermediate features are then dynamically reordered based on Y-coordinates and processed through a second Mamba block to model Y-direction relationships. 

Crucially, this architectural synergy eliminates the need for latency-heavy spatial sorting routines (e.g., space-filling curves). Because the interleaved LHA modules robustly anchor the local geometry throughout the network architecture, BDM is freed to efficiently deliver linear-complexity global perception without spatial truncation. Overall, the LAM block combines fine-grained local aggregation with global context modeling in an efficient way.

\section{Experiment}

\subsection{Implementation Details}
\label{sec:implementation}

Our framework is developed upon OpenPCDet~\cite{openpcdet2020}. For the LPS, the local spatial partitions are set to $0.3\text{m} \times 0.3\text{m} \times 0.25\text{m}$ for nuScenes and $0.32\text{m} \times 0.32\text{m} \times 0.1875\text{m}$ for Waymo, retaining the top-$k$ ($k=1$) salient points per region via DSS. The 3D backbone stacks $N=4$ PointLAM blocks.
Crucially, within each LHA module, we efficiently factorize the $5 \times 5 \times 5$ geometric receptive field into two consecutive $3 \times 3 \times 3$ sparse convolutions, preserving spatial context while decreasing the indexing operations per point.
Models are trained on NVIDIA A800 GPUs for 36 epochs on nuScenes and 24 epochs on Waymo. 
On nuScenes, epoch counts are not directly comparable across methods because Class-Balanced Grouping and Sampling (CBGS)~\cite{zhu2019class} changes the number of iterations per epoch. 
DSVT~\cite{wang2023dsvt}, HEDNet~\cite{zhang2023hednet}, and SAFDNet~\cite{zhang2024safdnet} use 20-epoch CBGS schedules, corresponding to 154,480 optimization steps according to their official implementations, whereas our 36-epoch schedule without CBGS contains 63,324 steps. 
Thus, 36 epochs does not imply a larger training budget. 
Except for disabling CBGS on nuScenes, we follow DSVT~\cite{wang2023dsvt} for optimization, data augmentation, and post-processing.

\begin{table}[t]
    \centering
    \caption{Performance comparison on the nuScenes dataset without using any test-time augmentation and model ensemble strategies. ‘C.V.’, ‘Ped.’, ‘M.C.’, and ‘T.C.’ denote construction vehicle, pedestrian, motorcycle, and traffic cone, respectively.}
    \setlength{\tabcolsep}{2.2mm}{}
    \resizebox{0.95\textwidth}{!}{
    \scalebox{1.0}{
        \begin{tabular}{lccccccccccccc}
        \toprule
        Method & Representation & NDS & mAP & Car & Truck & Bus & Trailer & C.V. & Ped. & M.C. & Bike & T.C. & Barrier \\ 
        \midrule
            \multicolumn{14}{c}{Results on the validation set} \\
        \midrule
        PillarNeXt~\cite{li2023pillarnext} & \multirow{2}{*}{Pillar} & 68.4 & 62.2 & 85.0 & 57.4 & 67.6 & 35.6 & 20.6 & 86.8 & 68.6 & 53.1 & {77.3} & 69.7 \\
        DSVT~\cite{wang2023dsvt} & & 71.1 & 66.4 & 87.4 & 62.6 & 75.9 & 42.1 & 25.3 & 88.2 & 74.8 & 58.7 & 77.8 & 70.9 \\
        \midrule
        CenterPoint~\cite{yin2021center} & \multirow{10}{*}{Voxel} & 66.5 & 59.2 & 84.9 & 57.4 & 70.7 & 38.1 & 16.9 & 85.1 & 59.0 & 42.0 & 69.8 & 68.3 \\
        TransFusion-L~\cite{bai2022transfusion} &  & 70.1 & 65.5 & 86.9 & 60.8 & 73.1 & 43.4 & 25.2 & 87.5 & 72.9 & 57.3 & 77.2 & 70.3 \\
        VoxelNeXt~\cite{chen2023voxelnext} &  & 66.7 & 60.5 & 83.9 & 55.5 & 70.5 & 38.1 & 21.1 & 84.6 & 62.8 & 50.0 & 69.4 & 69.4 \\
        HEDNet~\cite{zhang2023hednet} &  & 71.4 & 66.7 & 87.7 & 60.6 & 77.8 & 50.7 & 28.9 & 87.1 & 74.3 & 56.8 & 76.3 & 66.9\\
        FSDv2~\cite{fan2024fsd} &  & 70.4 & 64.7 & 83.7 & 51.6 & 66.4 & 59.1 & 32.5 & 87.1 & 71.4 & 51.7 & 80.3 & 78.7 \\
        SAFDNet~\cite{zhang2024safdnet} &  & 71.0 & 66.3 & 87.6 & 60.8 & 78.0 & 43.5 & 26.6 & 87.8 & 75.5 & 58.0 & 75.0 & 69.7\\
        LION~\cite{liu2024lion} &  & 72.1 & 68.0 & 87.9 & 64.9 & 77.6 & 44.4 & 28.5 & 89.6 & 75.6 & 59.4 & 80.8 & 71.6\\
        Voxel Mamba~\cite{zhang2024voxel} &  & 71.9 & 67.5 & 87.9 & 62.8 & 76.8 & 45.9 & 24.9 & 89.3 & 77.1 & 58.6 & 80.1 & 71.5\\  
        FSHNet~\cite{liu2025fshnet}&  &71.7 & 68.1 & 88.7& 61.4 & 79.3 & 47.8 & 26.3 & 89.3 & 76.7 & 60.5 & 78.6 & 72.3 \\
        UniMamba~\cite{jin2025unimamba}&  &72.6 & 68.5 & 88.7& 64.7 & 79.7 & 47.9 & 28.7 & 89.7 & 74.6 & 59.1 & 79.5 & 72.3 \\
        \midrule
        \textbf{PointLAM} (Ours) & Point & 72.2 & 67.8 & 88.6 & 64.1 & 78.9 & 45.3 & 26.5 & 89.2 & 73.9 & 59.5 & 80.2 & 72.1\\

        \midrule
            \multicolumn{14}{c}{Results on the test set} \\
        \midrule
                
        PointPillars~\cite{lang2019pointpillars} & \multirow{2}{*}{Pillar} &45.3 & 30.5 & 68.4 & 23.0 & 28.2 & 23.4 & 4.1 & 59.7 & 27.4 & 1.1 & 30.8 & 38.9\\
        PillarNet~\cite{shi2022pillarnet} &  &71.4 & 66.0 &  {87.6} & 57.5 & 63.6 & 63.1 & 27.9 & 87.3 & 70.1 & 42.3 & 83.3 & 77.2 \\
        
        \midrule
        
        CenterPoint~\cite{yin2021center} & \multirow{10}{*}{Voxel} &65.5 & 58.0 & 84.6 & 51.0 & 60.2 & 53.2 & 17.5 & 83.4 & 53.7 & 28.7 & 76.7 & 70.9\\
        VoxelNeXt~\cite{chen2023voxelnext} &  &70.0 & 64.5 & 84.6 & 53.0 & 64.7 & 55.8 & 28.7 & 85.8 & 73.2 & 45.7 & 79.0 & 74.6 \\
        TransFusion-L~\cite{bai2022transfusion} & & 70.2 & 65.5 & 86.2 & 56.7 & 66.3 & 58.8 & 28.2 & 86.1 & 68.3 & 44.2 & 82.0 & 78.2\\
        FSDv2~\cite{fan2024fsd} &  &71.7 & 66.2 & 83.7 & 51.6 & 66.4 & 59.1 & 32.5 & 87.1 & 71.4 & 51.7 & 80.3 & 78.7 \\
        LargeKernel3D~\cite{chen2023largekernel3d} &  &70.6 & 65.4 & 85.5 & 53.8 & 64.4 & 59.5 & 29.7 & 85.9 & 72.7 & 46.8 & 79.9 & 75.5\\
        LinK~\cite{lu2023link} &  &71.0 & 66.3 & 86.1 & 55.7 & 65.7 & 62.1 & 30.9 & 85.8 & 73.5 & 47.5 & 80.4 & 75.5\\
        HEDNet~\cite{zhang2023hednet} &  & 72.0 & 67.7 & 87.1 & 56.5 & 70.4 & 63.5 & 33.6 & 87.9 & 70.4 & 44.8 & 85.1 & 78.1 \\
        DSVT~\cite{wang2023dsvt} &  & 72.7 & 68.4 & 86.8 & 58.4 & 67.3 & 63.1 & 37.1 & 88.0 & 73.0 & 47.2 & 84.9 & 78.4 \\
        LION~\cite{liu2024lion} &  & 73.9 & 69.8 & 87.2 & 61.1 & 68.9 & 65.0 & 36.3 & 90.0 & 74.0 & 49.2 & 87.3 & 79.5 \\
        Voxel Mamba~\cite{zhang2024voxel} &  &73.0 & 69.0 & 86.8 & 57.1 & 68.0 & 63.2 & 35.4 & 89.5 & 74.7 & 50.8 & 86.9 & 77.3 \\
        UniMamba~\cite{jin2025unimamba}&  &74.0 & 70.2 & 87.9& 60.4 & 70.9 & 65.9 & 36.7 & 90.5 & 73.5 & 49.5 & 86.9 & 79.4 \\
        
        \midrule
        
        3DSSD~\cite{yang20203dssd} & \multirow{2}{*}{Point}  &56.4 & 42.6 & 81.2 & 47.2 & 61.4 & 30.5 & 12.6 & 70.2 & 36.0 & 8.6 & 31.1 & 47.9\\
        \textbf{PointLAM} (Ours) & & 73.0& 68.8& 87.8& 60.8& 72.4& 63.2& 25.7& 88.6& 71.4& 48.8& 86.3& 79.3 \\

        \bottomrule

        \end{tabular}}}
    \label{tab:main_nusc}
\end{table}

\begin{table}[t]
    \caption{Performance comparison on the Waymo Open Dataset (single-frame setting) without using any test-time augmentation and model ensemble strategies. Symbol `-' denotes that the result is not available.}
    \centering
    \resizebox{0.95\columnwidth}{!}{
        \begin{tabular}{lccccccccc}
        \toprule
         &  &\multicolumn{2}{c}{ALL (3D mAPH)} & \multicolumn{2}{c}{Vehicle (AP/APH)} & \multicolumn{2}{c}{Pedestrian (AP/APH)} & \multicolumn{2}{c}{Cyclist (AP/APH)} \\
        \multirow{-2}{*}{Method} & \multirow{-2}{*}{Representation} &L1 & L2 & L1 & L2 & L1 & L2 & L1 & L2 \\
            \midrule
                \multicolumn{10}{c}{Results on the validation set} \\
            \midrule
            
            PV-RCNN \cite{shi2020pv}& \multirow{2}{*}{Point-Voxel}&69.6& 63.3& 77.5 / 76.9& 69.0 / 68.4& 75.0 / 65.7& 66.0 / 57.6& 67.8 / 66.4& 65.4 / 64.0\\
            PV-RCNN++ \cite{shi2023pv} &  &75.2& 68.6& 79.1 / 78.6& 70.3 / 69.9& 80.6 / 74.6& 71.9 / 66.3& 73.5 / 72.4& 70.7 / 69.6\\
            
            \midrule

            PointPillar~\cite{lang2019pointpillars}& \multirow{6}{*}{Pillar} &63.3& 57.5& 71.6 / 71.0& 63.1 / 62.5& 70.6 / 56.7& 62.9 / 50.2& 64.4 / 62.3& 61.9 / 59.9\\
            PillarNet~\cite{shi2022pillarnet}&  &74.6&68.4&79.1 / 78.6& 70.9 / 70.5& 80.6 / 74.0& 72.3 / 66.2& 72.3 / 71.2& 69.7 / 68.7\\
            SWFormer~\cite{sun2022swformer} &  &-& -&77.8 / 77.3& 69.2 / 68.8&80.9 / 72.7 &72.5 / 64.9 &-&- \\
            PillarNeXt~\cite{li2023pillarnext} &  &75.7& 69.7&78.4 / 77.9& 70.3 / 69.8&82.5 / 77.1& 74.9 / 69.8 &73.2 / 72.2&70.6 / 69.6 \\
            FlatFormer~\cite{liu2023flatformer} &   &- & 67.2 & - & 69.0 / 68.6 & - & 71.5 / 65.3 & -  & 68.6 / 67.5 \\
            PTv3~\cite{wu2024pointv3} &   &- & 70.5 & -  & 71.2 / 70.8 & - & 76.3 / 70.4 & - & 71.5 / 70.4 \\

            \midrule

            SECOND~\cite{yan2018second} &  \multirow{10}{*}{Voxel} &63.1& 57.2 &72.3 / 71.7& 63.9 / 63.3& 68.7 / 58.2& 60.7 / 51.3& 60.6 / 59.3& 58.3 / 57.1 \\
            Centerpoint~\cite{yin2021center} &  &73.5 & 67.6 & 76.6 / 76.0 & 68.9 / 68.4& 79.0 / 73.4 & 71.0 / 65.8 & 72.1 / 71.0 & 69.5 / 68.5\\
            VoxelNeXt~\cite{chen2023voxelnext} &  &76.3 & 70.1 & 78.2 / 77.7 & 69.9 / 69.4 &81.5 / 76.3 &73.5 / 68.6 &76.1 / 74.9 &73.3 / 72.2\\
            DSVT~\cite{wang2023dsvt}&  &78.2 & 72.1 & 79.7 / 79.3& 71.4 / 71.0& 83.7 /     78.9& 76.1 / 71.5& {77.5} / 76.5& 74.6 / 73.7\\
            HEDNet~\cite{zhang2023hednet} &  &79.4 & 73.4 & 81.1 / 80.6 & 73.2 / 72.7 & 84.4 / 80.0 & 76.8 / 72.6 & 78.7 / 77.7 & 75.8 / 74.9\\
            SAFDNet~\cite{zhang2024safdnet} &   &79.2 & 73.2 & 80.2 / 79.7 & 72.2 / 71.8 & 79.9 / 76.9 & 76.8 / 72.6 & 79.1 / 78.1 & 76.2 / 75.2 \\
            LION~\cite{liu2024lion}&  &80.1 & 74.0 & 80.3 / 79.9 & 72.0 / 71.6 & 85.8 / 81.4 & 78.5 / 74.3 & 80.1 / 79.0 & 77.2 / 76.2 \\
            Voxel Mamba~\cite{zhang2024voxel} &  &79.6 & 73.6 & {80.8} / {80.3}& {72.6} / {72.2}& {85.0} / {80.8}& {77.7} / {73.6}& {78.6} / {77.6}& {75.7} / {74.8}\\
            UniMamba~\cite{jin2025unimamba}&  &80.2 & 74.1 & 80.6 / 80.0& 72.3 / 71.8& 86.0 / 81.3& 78.7 / 74.1& 80.3 / 79.3& 77.5 / 76.5\\
            FSHNet~\cite{liu2025fshnet}&  & 80.6& 74.9 & 82.2 / 81.7& 74.5 / 74.0& 85.9 / 80.8& 78.9 / 73.9& 80.5 / 79.4& 78.0 / 76.9\\

            \midrule

            \textbf{PointLAM} (Ours) & Point & 79.7 & 73.6 & 80.0 / 79.6 & 71.7 / 71.3 & 85.1 / 81.1 & 78.0 / 74.1 & 79.4 / 78.3 & 76.5 / 75.5 \\

            \midrule
                \multicolumn{10}{c}{Results on the test set} \\
            \midrule

                PV-RCNN~\cite{shi2020pv} &\multirow{2}{*}{Point-Voxel}&  74.2  & 68.8 & 80.6 / 80.1 & 72.8 / 72.4 & 78.2 / 72.0 & 71.8 / 66.0 & 71.8 / 70.4 & 69.1 / 67.8 \\
				PV-RCNN++~\cite{shi2023pv}  & & 75.7& 70.2&81.6 / 81.2& 73.9 / 73.5& 80.4 / 75.0& 74.1 / 69.0& 71.9 / 70.8& 69.3 / 68.2 \\
                
            \midrule
				PointPillar~\cite{lang2019pointpillars} &\multirow{2}{*}{Pillar}&-&-& 68.6 / 68.1& 60.5 / 60.1& 68.0 / 55.5& 61.4 / 50.1&-&-\\
                PillarNeXt-3f~\cite{li2023pillarnext}  &&79.0 & 74.1 & 83.3 / 82.8 & 76.2 / 75.8 & 84.4 / 81.4	& 78.8 / 76.0 & 73.8 / 72.7	& 71.6 / 70.6\\
            \midrule
				CenterPoint~\cite{yin2021center} &\multirow{7}{*}{Voxel}& 77.2& 71.9& 81.1 / 80.6 &73.4 / 73.0& 80.5 / 77.3& 74.6 / 71.5& 74.6 / 73.7& 72.2 / 71.3 \\
                AFDetV2~\cite{hu2022afdetv2} && 75.2 & 70.3 & 80.5 / 80.0 & 73.0 / 72.6 & 79.8 / 74.3 & 73.7 / 68.6 & 72.4 / 71.2 & 69.8 / 69.7 \\
				SST-3f~\cite{fan2022embracing}  &&78.3 &72.8 &81.0 / 80.6 &73.1 / 72.7& 83.3 / 79.7 &76.9 / 73.5 &{75.7} / {74.6} &{73.2} / {72.2} \\
                FSDv1~\cite{fan2022fully}  && 78.2 & 72.4 & 82.7 / 82.3 & 74.4 / 74.1 & 82.9 / 77.9 & 75.9 / 71.3 & 75.6 / 74.4 & 72.9 / 71.8 \\
                FSDv2~\cite{fan2024fsd}  && 79.0 & 73.3 & 82.4 / 82.0 & 74.4 / 74.0 & 83.8 / 78.9 & 77.4 / 72.8 & 77.1 / 76.0 & 74.3 / 73.2\\
                SAFDNet~\cite{zhang2024safdnet} && 79.8 & {74.6} & 83.9 / 83.5 & 76.6 / 76.2 & 84.3 / 79.8 & 78.4 / 74.1 & 77.5 / 76.3 & 74.6 / 73.4 \\
                Voxel Mamba~\cite{zhang2024voxel} &&{79.6}&{74.3}& {84.4} / {84.0} &{77.0} / {76.6} & {84.8} / {80.6} &{79.0} / {74.9}&{75.4} / {74.3}& {72.6} / {71.5}\\
                FSHNet~\cite{liu2025fshnet}&  & 80.3& 75.2 & 84.9 / 84.5& 77.8 / 77.4& 85.3 / 80.0& 79.5 / 74.3& 77.7 / 76.5& 74.8 / 73.7\\
            \midrule
                
                \textbf{PointLAM} (Ours)  &\multirow{1}{*}{Point}&{79.8}&{74.4}& {83.2} / {82.9} &{75.5} / {75.2} & {85.2} / {80.7} &{79.4} / {75.2}&{76.9} / {75.7}& {74.1} / {72.9}\\

            \bottomrule
        \end{tabular}}
	\label{tab:main_waymo}
\end{table}

\subsection{Main Results}

\noindent\textbf{nuScenes.}
As shown in Table~\ref{tab:main_nusc}, PointLAM achieves competitive performance on the nuScenes dataset while maintaining a point-based architecture. On the validation set, PointLAM yields 72.2 NDS and 67.8 mAP, surpassing prominent voxel-based models such as DSVT (71.1 NDS) and LION (72.1 NDS), and remaining comparable to the recent UniMamba (72.6 NDS). On the test set, PointLAM establishes a strong baseline for point-based detectors with 73.0 NDS and 68.8 mAP, significantly outperforming the previous point-based 3DSSD (+16.6 NDS). It also exactly matches the performance of the highly optimized voxel-based framework Voxel Mamba (73.0 NDS) and outperforms DSVT (72.7 NDS). These results validate the efficacy of our proposed point-level operators in successfully bridging the performance gap with state-of-the-art voxel methods.

\noindent\textbf{Waymo.}
As shown in Table~\ref{tab:main_waymo}, PointLAM demonstrates competitive performance on the Waymo Open Dataset while adhering to a point-based architecture. On the validation set, it achieves 73.6 L2 mAPH, outperforming established detectors such as DSVT (72.1 L2) and HEDNet (73.4 L2), and matching the strong Voxel Mamba baseline. Similarly, on the official test benchmark, PointLAM yields 79.8 L1 mAPH and 74.4 L2 mAPH, performing comparably to the state-of-the-art voxel-based SAFDNet (74.6 L2). Notably, without relying on dense volumetric representations, our model exhibits robust capabilities on geometrically delicate classes, yielding highly competitive scores on pedestrians (e.g., 75.2 L2 APH on the test set). This intrinsic strength empirically reflects the advantage of point-based paradigms in preserving fine-grained structural fidelity free from quantization loss.

\begin{table}[t]
\centering
\caption{Comparison of parameter count, computation cost, and inference latency of different 3D detection backbones.}
\label{tab:efficiency}
\resizebox{\textwidth}{!}{
\begin{tabular}{ccccccccc}
\toprule
Method & Venue &  Backbone&Representation & \#Parameters (M) & FLOPs (G) & Latency (ms) & NDS &L2 mAPH\\
\midrule
PVRCNN~\cite{shi2020pv} & CVPR'20  & \multirow{8}{*}{spCNN} &Point-Voxel  & 8.5 & 47.4 &281.2& - &63.3\\
CenterPoint~\cite{yin2021center} & CVPR'21 & &Voxel  & 2.7 & 41.8 &60.1& 66.5 &67.6\\
VoxelRCNN~\cite{deng2021voxel} & AAAI'21 & &Voxel  & 11.7 & 23.3 &94.5& - &66.2\\
VoxelNeXt~\cite{chen2023voxelnext} & CVPR'23 & &Voxel  & 15.5 & 97.5 &192.9& 66.7 &70.1\\
PVRCNN++~\cite{shi2023pv} & IJCV'23 & &Point-Voxel  &9.5 & 55.4&97.3& - &68.6\\
HEDNet~\cite{zhang2023hednet} & NeurIPS'23 & &Voxel & 4.6& 106.2& 96.7 & 71.4 &73.4\\
SAFDNet~\cite{zhang2024safdnet} & CVPR'24 & &Voxel & 3.7& 60.8&104.3& 71.0 &73.2\\
FSHNet~\cite{liu2025fshnet} & CVPR'25 & &Voxel & 10.7& 161.6&99.2& 71.7 &74.9\\
\midrule

FlatFormer~\cite{liu2023flatformer} & CVPR'23 & \multirow{3}{*}{Transformer}  &Pillar & 1.1 & 48.1 & 62.3& - & 67.2\\
DSVT-Pillar~\cite{wang2023dsvt} & CVPR'23 &  & Pillar & 1.2 & 35.6 &73.2& 71.1 &71.0\\
DSVT-voxel~\cite{wang2023dsvt} & CVPR'23 &  & Voxel & 2.7 & 108.6 &115.6& - &72.1\\
\midrule

Voxel Mamba~\cite{zhang2024voxel} & NeurIPS'24 & \multirow{3}{*}{Mamba} & Voxel & 15.1 & 246.2 &109.8& 71.9 &73.6\\ 
LION~\cite{liu2024lion} & NeurIPS'24 & &Voxel & 10.1 & 165.8 &195.3& 72.1 &74.0\\ 
\textbf{PointLAM} (Ours) & - & &Point & 8.6 & 90.7& 93.1& 72.2 & 73.6 \\
\bottomrule
\end{tabular}}
\end{table}

\noindent\textbf{Efficiency.}
Table~\ref{tab:efficiency} benchmarks PointLAM's computational footprint and latency on a single NVIDIA A800 GPU.
Yielding top-tier accuracy (72.2 NDS, 73.6 L2 mAPH), our point-based framework operates with 8.6M parameters, 90.7G FLOPs, and 93.1ms latency, establishing a highly competitive trade-off against established voxel-based methods. Specifically, while HEDNet~\cite{zhang2023hednet} maintains a more compact parameter footprint (4.6M), PointLAM delivers superior precision (+0.5 NDS, +0.2 L2 mAPH) with fewer FLOPs (90.7G vs. 106.2G) and lower latency (93.1ms vs. 96.7ms). Compared to LION~\cite{liu2024lion}, PointLAM achieves higher nuScenes accuracy while slashing FLOPs by 45\% (90.7G vs. 165.8G) and accelerating inference by over 2$\times$ (93.1ms vs. 195.3ms). Furthermore, it matches Voxel Mamba's~\cite{zhang2024voxel} Waymo performance utilizing 37\% of its FLOPs.
These striking margins validate that replacing heuristic sampling and density-sensitive spatial queries with our LPS and LHA routing intrinsically resolves the systemic bottlenecks of the point-based paradigm, efficiently mitigating the accuracy-efficiency trade-off while retaining point-level features.

\noindent\textbf{Robustness.}
To systematically evaluate model robustness, we analyze detection performance across varying global input densities, instance-level point counts, and physical object sizes (Tables~\ref{tab:abl_density}~\ref{tab:abl_pointnum}~\ref{tab:abl_size}). Under uniform density reduction (down to 1/16) and extreme instance-level sparsity (ranging from 5 to 25 points per object), PointLAM consistently outperforms the highly optimized voxel baseline, LION. These results validate that our point-based architecture intrinsically preserves critical fine-grained geometries even when spatial information is severely degraded. Furthermore, we examine the efficacy of our Local Hadamard Aggregator (LHA) across different physical sizes by comparing it against a $k$-NN baseline.  While $k$-NN enforces a fixed neighbor count—inevitably retrieving distant background noise when targets are small or sparsely populated—our transient grid imposes a strictly bounded spatial receptive field. This deterministic routing structurally isolates feature contamination, yielding the most substantial performance advantages on the smallest instances (e.g., $+1.24$ L1 mAPH over $k$-NN at Size $1.0m^3$). This directly corroborates the theoretical superiority of bounded discrete indexing over dynamic continuous queries in preserving pure object features.

\begin{table*}[t]
  \centering
  \begin{minipage}[t]{0.48\linewidth}
    \centering
    \setlength{\tabcolsep}{1.9mm}
    \caption{Robustness comparison under varying input point densities.} \label{tab:abl_density}
    \resizebox{0.9\columnwidth}{!}{

    \begin{tabular}{ccccc}
\toprule
\multirow{2}{*}{Density}& \multicolumn{2}{c}{Ours}& \multicolumn{2}{c}{LION~\cite{liu2024lion}} \\
\cmidrule(lr){2-3} \cmidrule(lr){4-5}
 & L1 mAPH & L2 mAPH & L1 mAPH & L2 mAPH  \\
\midrule
1/2 & \textbf{70.06} & \textbf{63.78} &69.47 &63.23  \\
1/3 & \textbf{62.28} & \textbf{56.17} &61.52 &55.44  \\
1/4 & \textbf{56.70} & \textbf{50.82} &55.37 &49.55  \\
1/8 & \textbf{40.52} & \textbf{35.80} &39.23 &34.57  \\
1/16 & \textbf{25.08} & \textbf{21.86} &23.72 &20.60  \\
\bottomrule
\end{tabular}

} 
\end{minipage}\quad
  \begin{minipage}[t]{0.48\linewidth}
    \centering
    \setlength{\tabcolsep}{1.8mm}
    \caption{Robustness comparison across different points number in one object.} \label{tab:abl_pointnum}
    \resizebox{0.9\columnwidth}{!}{

\begin{tabular}{ccccc}
\toprule
\multirow{2}{*}{Number}& \multicolumn{2}{c}{Ours}& \multicolumn{2}{c}{LION~\cite{liu2024lion}} \\
\cmidrule(lr){2-3} \cmidrule(lr){4-5}
& L1 mAPH & L2 mAPH & L1 mAPH & L2 mAPH  \\
\midrule
5& \textbf{21.40} & \textbf{19.76} &19.98 &18.39 \\
10& \textbf{22.50} & \textbf{20.71} &21.46 &19.71\\
15& \textbf{24.33} & \textbf{22.36} & 23.00 &21.07 \\
20& \textbf{26.27} & \textbf{24.10} &24.76 &22.63\\
25& \textbf{28.50} & \textbf{26.13} &26.84 &24.52\\
\bottomrule
\end{tabular}

}
\end{minipage}
\end{table*}

\begin{table}[t]
    \centering
    \caption{Robustness comparison across different object sizes.}
    \label{tab:abl_size}
    \resizebox{0.6\columnwidth}{!}{
\begin{tabular}{ccccccc}
\toprule
\multirow{2}{*}{Size ($m^{3}$)}& \multicolumn{2}{c}{Ours} & \multicolumn{2}{c}{LION~\cite{liu2024lion}}& \multicolumn{2}{c}{Ours-KNN} \\
\cmidrule(lr){2-3} \cmidrule(lr){4-5} \cmidrule(lr){6-7}
& L1 mAPH & L2 mAPH  & L1 mAPH & L2 mAPH & L1 mAPH & L2 mAPH  \\
\midrule
1.0& \textbf{3.75} & \textbf{3.28}  & 3.38 &2.93 &2.51 &2.17 \\
1.5& \textbf{20.70} & \textbf{18.82}  & 20.26 & 18.37 &19.09 &17.27 \\
2.0& \textbf{29.75} & \textbf{27.27}  & 29.36 & 26.85 &27.77 &25.34 \\
2.5& \textbf{35.40} & \textbf{32.69}  & 34.95 & 32.23 &33.76 &31.08 \\
3.0& \textbf{47.12} & \textbf{43.97}  & 46.12 & 42.97 &43.72 &40.68 \\
\bottomrule
\end{tabular}}

\end{table}


\setlength{\intextsep}{1pt}
\begin{wrapfigure}[18]{r}{0.4\textwidth}
    \centering
    \includegraphics[width=\linewidth]{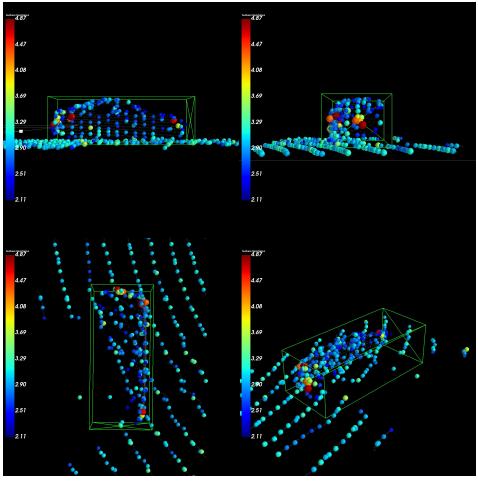}
    \caption{Visualization of LPS sampling and LHA feature modulation. Raw input points are in light gray. Points retained by LPS are colored based on their LHA activation magnitude, highlighting high-frequency structural boundaries.}
    \label{fig:visual}
\end{wrapfigure}

\noindent\textbf{Interpretability of LPS and LHA.}  
Figure~\ref{fig:visual} visualizes our sampling behavior and feature activations. It explicitly demonstrates the high-pass filtering nature of our Laplacian Point Sampler (LPS): it aggressively discards low-frequency flat surfaces while preserving the informative geometric skeletons of foreground objects. Furthermore, the heatmaps reveal that the Hadamard modulation in LHA is highly topology-aware. High activations align perfectly with distinct structural boundaries and corners, whereas responses on planar regions remain suppressed, directly validating our geometric prior design.

\subsection{Ablation Studies}

We adopt an ablation strategy driven by the distinct physical characteristics of two datasets. Experiments on the highly sparse nuScenes dataset (18-epoch schedule) evaluate feature representation capacity and module synergy. Conversely, the $1/5$ subset of the massive, dense Waymo Open Dataset serves as a stress test for operator scalability, computational efficiency, and spatial indexing.


\begin{table}[t]
  \centering
  \begin{minipage}[t]{0.48\linewidth}
    \centering
    \setlength{\tabcolsep}{1.9mm}
    \caption{Ablation on Laplacian Point Sampler components.} \label{tab:abl_dps}
    \resizebox{0.8\columnwidth}{!}{

\begin{tabular}{ccccc}
\toprule
Encoding & Sampling & mAP &   NDS \\
\midrule
PFN & Pooling    &  67.41 & 71.42   \\
DevNet & Pooling &   67.82  & 71.59   \\
PFN & DSS     &  67.77 & 71.56   \\
DevNet & DSS      &  \textbf{68.14} & \textbf{71.82} \\
\bottomrule 
\end{tabular}

} 
\end{minipage}
\hfill
  \begin{minipage}[t]{0.48\linewidth}
     \centering
     \setlength{\tabcolsep}{2.7mm}
     \caption{Effect of aggregation modes in the Local Hadamard Aggregator.} \label{tab:abl_aggre}
     \resizebox{0.8\columnwidth}{!}{
     
\begin{tabular}{ccc}
\toprule
Aggregation Mode & mAP & NDS   \\
\midrule
Addition & 67.13& 70.98  \\
Subtraction & 67.51& 71.53 \\
Hadamard & \textbf{68.14} & \textbf{71.82}  \\
Concatenation & 67.63 & 71.51  \\
\bottomrule
\end{tabular}

} 
\end{minipage}
\end{table}

\noindent\textbf{Effectiveness of Laplacian Point Sampler.}
Table~\ref{tab:abl_dps} dissects the LPS into its encoding and sampling phases. A standard voxelization-like baseline using a Point Feature Network (PFN) and local pooling yields 71.42 NDS. Upgrading the encoder to our DevNet improves performance to 71.59 NDS, proving that explicitly capturing high-frequency structural boundaries is superior to semantic-agnostic aggregation. Concurrently, replacing pooling with our Doubly Sorted Sampling (DSS) achieves 71.56 NDS. Synergizing both delivers the optimal 71.82 NDS. This confirms that in highly sparse regimes, encoding geometric priors and preserving structural skeletons are equally indispensable.

\noindent\textbf{Effectiveness of Aggregation Mode in LHA.}
Table~\ref{tab:abl_aggre} evaluates the aggregation mechanisms within LHA. In sparse conditions without explicit continuous positional embeddings, simple linear operations like addition (70.98 NDS) or concatenation (71.51 NDS) fail to achieve sufficient channel-wise interaction. In contrast, our Hadamard product achieves a peak 71.82 NDS, outperforming the next-best mode by +0.29 NDS. This validates that a Hadamard gating mechanism acts as a highly expressive, dynamic feature-wise filter, adaptively modulating local topology with minimal computational overhead.

\begin{table*}[t]
  \centering
  \begin{minipage}[t]{0.46\linewidth}
    \centering
    \setlength{\tabcolsep}{1.9mm}
    \caption{Ablation on architectural synergy between LHA and BDM.} \label{tab:abl_backbone}
    \resizebox{0.9\columnwidth}{!}{

    \begin{tabular}{ccccc}
    \toprule
     \multirow{2}{*}{LHA} & \multicolumn{2}{c}{BDM }   
    &\multirow{2}{*}{mAP} & \multirow{2}{*}{NDS}  \\
    \cmidrule(lr){2-3}
      & {\small X-axis} & {\small Y-axis} &   &    \\
    \midrule
     \checkmark& &   & 64.37&69.40 \\
      &\checkmark &  &63.67& 68.68  \\
      & \checkmark & \checkmark  &64.87& 69.58  \\
     \checkmark & \checkmark  &  &67.26& 71.23  \\
     \checkmark & \checkmark &  \checkmark &\textbf{68.14}& \textbf{71.82} \\
    \bottomrule 
    \end{tabular}

} 
\end{minipage}\quad
  \begin{minipage}[t]{0.50\linewidth}
    \centering
    \setlength{\tabcolsep}{1.8mm}
    \caption{Impact of Mamba scan orders on performance and latency. } \label{tab:abl_scan}
    \resizebox{0.8\columnwidth}{!}{

    \begin{tabular}{cccc}
\toprule
         Scan order &  mAP &  NDS & Latency (ms)\\
\midrule
         Random &  66.74 &  71.08 & 0.1\\
         Hilbert &  67.06 &  \textbf{71.26} &5.5\\
         Z-order &  \textbf{67.27} &  71.23 &5.6\\
         Shuffle &  66.35 &  70.62 & 11.2\\
         Shuffle$^{*}$ &  66.75 &  70.89 &11.2\\
         Axis Sort &  \underline{67.26}&  \underline{71.23}&0.1\\
\bottomrule
    \end{tabular}

} 
\end{minipage}
\end{table*}

\noindent\textbf{Architectural Synergy of Backbone Modules.}
Table~\ref{tab:abl_backbone} demonstrates the indispensable, highly complementary relationship between LHA and Bi-Directional Mamba (BDM). Relying solely on serialized BDM yields a sub-optimal 69.58 NDS, as pure sequence models inherently struggle with unstructured sparse points and easily overfit to serialization artifacts. Conversely, utilizing LHA alone achieves only 69.40 NDS due to the absence of global receptive fields. Synergizing both boosts performance by a massive +2.24 NDS (to 71.82). This confirms our structural paradigm: LHA's robust local topological extraction perfectly mitigates the spatial information loss of 1D serialization, empowering BDM to efficiently establish global dependencies.

\noindent\textbf{Effectiveness of Mamba Scan Orders.}
Table~\ref{tab:abl_scan} compares Axis Sort against random sequences, space-filling curves (``Hilbert'', ``Z-order''), and their block-wise mixtures (``Shuffle'', ``Shuffle$^{*}$''). While complex curves yield marginal gains by preserving spatial locality, they incur a prohibitive 50$\times$ latency overhead. Intriguingly, even the geometrically agnostic ``Random'' scan remains highly competitive. This robustness empirically validates our architectural synergy: because LHA modules explicitly anchor local topologies prior to serialization, BDM's global perception is fundamentally desensitized to precise sequence ordering. The simple Axis Sort delivers the optimal precision-efficiency trade-off.


\begin{table*}[t]
  \centering
  \begin{minipage}[t]{0.46\linewidth}
    \centering
    \setlength{\tabcolsep}{1.9mm}
    \caption{Validation of Laplacian geometric priors and DSS sensitivity to $k$.} \label{tab:importance_scoring}
    \resizebox{0.9\columnwidth}{!}{

    \begin{tabular}{ccc}
        \toprule
        Method & L1 mAPH & L2 mAPH\\
        \midrule
        DSS-Random (k=1.0) & 74.59 & 68.50 \\
        \textbf{DSS-IS (k=1.0)} & \textbf{75.02} & \textbf{68.90} \\
        DSS-Random (k=1.5) & 74.57 & 68.48 \\
        \textbf{DSS-IS (k=1.5)} & \textbf{74.99} & \textbf{68.91} \\
        DSS-Random (k=2.0) & 74.59 & 68.46 \\
        \textbf{DSS-IS (k=2.0)} & \textbf{75.14} & \textbf{69.02} \\
        DSS-Random (k=2.5) & 74.51 & 68.39 \\
        \textbf{DSS-IS (k=2.5)} & \textbf{74.93} & \textbf{68.79} \\
        \bottomrule
    \end{tabular}

} 
\end{minipage}\quad
  \begin{minipage}[t]{0.50\linewidth}
    \centering
    \setlength{\tabcolsep}{1.8mm}
    \caption{Ablation on neighbor queries and kernel sizes in LHA.} \label{tab:abl_LHA}
    \resizebox{0.9\columnwidth}{!}{
    
\begin{tabular}{cccc}
\toprule
Neighbor & L1 mAPH & L2 mAPH  &Latency (ms) \\
\midrule
KNN& 74.61 &  68.40 &44.5   \\
PTv3 & 72.52  & 66.53  &11.2  \\
Ours & \textbf{75.95}  & \textbf{69.87} &\textbf{2.7}   \\
\midrule
Kernel & L1 mAPH & L2 mAPH  &Latency (ms) \\
\midrule
$3\times3\times3$ & 75.63 & 69.54  &80.7  \\
$5\times5\times5$ & 75.95  & 69.87 &93.1 \\
$7\times7\times7$ & 76.02  & 69.95 &104.2  \\
\bottomrule
\end{tabular}

} 
\end{minipage}
\end{table*}

\noindent\textbf{Validation of Laplacian Geometric Priors.}
To efficiently explore the hyperparameter space, we conduct this ablation on a 10\% Waymo subset. Table~\ref{tab:importance_scoring} validates our Laplacian saliency scores across varying sampling densities ($k$). Replacing our Laplacian-guided sorting (DSS-IS) with random selection (DSS-Random) consistently degrades performance across all $k$. This empirically confirms that our implicit Laplacian filter effectively isolates and anchors the high-frequency structural skeletons of foreground objects. Moreover, the sustained performance stability across diverse $k$ thresholds underscores the robust, plug-and-play nature of this geometric prior.

\noindent\textbf{Efficiency of LHA Neighbor Query.}
Table~\ref{tab:abl_LHA} (top) benchmarks our transient grid-based routing against traditional spatial queries. The continuous spatial search of $k$-NN is fundamentally density-sensitive, resulting in a prohibitively expensive latency (44.5 ms). While PTv3's serialization mitigates this, computing space-filling curves and reordering points still incurs a heavy 11.2 ms overhead. In stark contrast, our transient grid acts as a deterministic $O(1)$ spatial router, slashing routing latency to a negligible 2.7 ms. Crucially, this grid imposes a strict physical boundary that inherently isolates distant noise in dense regions, yielding the peak accuracy of 75.95 L1 mAPH.

\noindent\textbf{Effectiveness of Kernel Size in LHA.}
Table~\ref{tab:abl_LHA} (bottom) investigates the spatial receptive field of LHA by varying kernel sizes. While a compact $3\times3\times3$ kernel is highly efficient (80.7 ms), it confines the local context. Expanding to $5\times5\times5$ captures a broader geometric neighborhood, providing essential structural robustness against varying point densities and yielding a +0.32 L1 mAPH improvement (93.1 ms). However, further enlarging to $7\times7\times7$ triggers severe diminishing returns, gaining a mere +0.07 mAPH for an additional 11.1 ms overhead. We therefore adopt the $5\times5\times5$ kernel as our default, deliberately securing stable, robust geometric perception while successfully averting the severe latency bloat characteristic of unconstrained neighborhood queries.

\section{Conclusion}

In this work, we present PointLAM, a highly efficient point-based 3D object detector that mitigates the trade-off between geometric fidelity and computational cost. To overcome inherent point-based bottlenecks, we introduce two synergistic designs: the Laplacian Point Sampler (LPS) for fast, structure-aware downsampling using an implicit Laplacian high-pass filter, and the Local Hadamard Aggregator (LHA) for local spatial indexing and parameter-free topology modulation. Crucially, by structurally coupling LHA with Bi-Directional Mamba (BDM) layers, we formulate the Local Attentive Mamba (LAM) block. This core building block interleaves fine-grained local attentive gating with linear-complexity global context perception.
PointLAM achieves competitive performance on nuScenes and Waymo while maintaining a lightweight point-based design. It approaches or matches several strong voxel-based baselines with fewer parameters and lower latency. We hope PointLAM encourages point-level designs and inspires future research in efficient 3D perception.

\section*{Acknowledgements}

This work was supported by the National Natural Science Foundation of China (Grant Nos. 62322113, 62376156), as well as the Shanghai Municipal Special Program for Basic Research on General AI Foundation Models (Grant No. 2025SHZDZX025G15).

\newpage
%
%
\bibliographystyle{splncs04}
\bibliography{main}

\end{document}